\documentclass{article}

\usepackage{epsfig}
\usepackage{graphicx}
\usepackage{bm}
\usepackage{amssymb}
\usepackage{amsthm}
\usepackage{amsmath}
 \usepackage{multicol}
\usepackage{enumerate}
\usepackage{algorithm}
\usepackage{algorithmic}
\usepackage{multirow}
\DeclareMathOperator*{\argmin}{argmin}

\def\*#1{\mathbf{#1}}
\def\##1{\mathcal{#1}}

\newtheorem{theorem}{Theorem}

\newtheorem{assumption}{Assumption}

\usepackage{hyperref}

\usepackage{microtype}
\usepackage{subfigure}
\usepackage{booktabs} 
\usepackage{color}

\usepackage[accepted]{icml2019_edit}

\newtheorem{lem}{Lemma}[section]

\icmltitlerunning{Differentiable Linearized ADMM}

\begin{document}

\twocolumn[
\icmltitle{Differentiable Linearized ADMM}



\icmlsetsymbol{equal}{*}

\begin{icmlauthorlist}
\icmlauthor{Xingyu Xie}{equal,to}
\icmlauthor{Jianlong Wu}{equal,to}
\icmlauthor{Zhisheng Zhong}{to}
\icmlauthor{Guangcan Liu}{ed}
\icmlauthor{Zhouchen Lin}{to}
\end{icmlauthorlist}

\icmlaffiliation{to}{Key Lab. of Machine Perception, School of EECS, Peking University.}
\icmlaffiliation{ed}{B-DAT and CICAEET, School of Automation, Nanjing University of Information Science and Technology.}

\icmlcorrespondingauthor{Guangcan Liu}{gcliu@nuist.edu.cn}
\icmlcorrespondingauthor{Zhouchen Lin}{zlin@pku.edu.cn}

\icmlkeywords{Machine Learning, ICML}

\vskip 0.3in
]



\printAffiliationsAndNotice{\icmlEqualContribution} 

\begin{abstract}
Recently, a number of learning-based optimization methods that combine data-driven architectures with the classical optimization algorithms have been proposed and explored, showing superior empirical performance in solving various ill-posed inverse problems, but there is still a scarcity of rigorous analysis about the convergence behaviors of learning-based optimization. In particular, most existing analyses are specific to unconstrained problems but cannot apply to the more general cases where some variables of interest are subject to certain constraints. In this paper, we propose Differentiable Linearized ADMM (D-LADMM) for solving the problems with linear constraints. Specifically, D-LADMM is a $K$-layer LADMM inspired deep neural network, which is obtained by firstly introducing some learnable weights in the classical Linearized ADMM algorithm and then generalizing the proximal operator to some learnable activation function. Notably, we rigorously prove that there exist a set of learnable parameters for D-LADMM to generate globally converged solutions, and we show that those desired parameters can be attained by training D-LADMM in a proper way. To the best of our knowledge, we are the first to provide the convergence analysis for the learning-based optimization method on constrained problems.
\end{abstract}
\section{Introduction}
\label{sec:introduction}
Numerous problems solving at the core of statistics, learning and vision areas rely on well-designed optimization algorithms, and especially so for the recently prevalent deep learning. Provided with some well-deigned optimization strategies such as~\cite{kingma2014adam,zeiler2012adadelta,li2018lifted}, the researchers can focus on the design of task-oriented loss functions without being encumbered by the solving methods. In addition, optimization can also help the deep neural network (DNN) design. For example,~\citet{li2018optimization} show that optimization algorithms can in fact inspire the architectures of DNN, and they connect the classical optimization algorithms with some prevalent DNN architectures, e.g., ResNet~\cite{he2016deep} and DenseNet~\cite{huang2017densely}. While it is apparent that optimization does benefit learning, the converse of the statement is not so affirmative. That is, can the well-developed learning methods also benefit the optimization? If so, in what sense?

To answer the highlighted question, some techniques have been proposed to combine data-driven learning frameworks with the traditional optimization algorithms, so called as \emph{learning-based optimization}~\cite{gregor2010learning,liu2016learning,chen2017trainable,liu2018reversed,peng2018k}. Usually, the combination is achieved by introducing learnable parameters into the classical numerical solvers at first, then performing discriminative learning on collected training data so as to obtain some task-specific (but possibly inconsistent) optimization schemes. Due to the success of deep learning in a wide variety of application fields, many researchers choose to consider DNN as the learnable units for being combined with the optimization procedure. For example,~\citet{sprechmann2015learning,liu2018alista,chen2018theoretical} resemble a recurrent neural network (RNN) to solve the LASSO problem, and~\citet{zhou2018sc2net} show the connection between sparse coding and long short term memory~(LSTM). The empirical results in these studies illustrate that the computational efficiency of optimization is dramatically improved by the incorporation of DNN. However, there is only few work that analyzes the convergence properties of these algorithms in theory.~\citet{chen2018theoretical} prove that there exist a sequence of parameters for their learning-based optimization procedure to converge linearly to the optimal solution of the LASSO problem. But this result is specific to LASSO and may not apply to the other problems.

While most existing methods and theories in learning-based optimization are made specific to unconstrained problems, many optimization problems arising from modern applications may contain some constraints. In this paper, we would like to take a step towards learning-based \textit{constrained} optimization. To be more precise, we shall consider the following linearly constrained problem:
\begin{equation}\label{Gene:L-ADMM}
\min_{\*{Z}, \*{E}}~  f(\*{Z}) +  g(\*{E}), ~~\textrm{s.t.} \ \*{X}=\*{A}\*{Z}+\*B\*{E},
\end{equation}
where $\*{A}\in \mathbb{R}^{m\times d_1}, \*B \in \mathbb{R}^{m\times d_2}$, $\*{X} \in \mathbb{R}^{m\times n}$, and $f(\cdot)$ and $g(\cdot)$ are convex functions. Many problems in the learning field can be formulated as (\ref{Gene:L-ADMM}), e.g., matrix recovery~\cite{zhang2018primal,zhang2015exact,liu2016low,liu2017blessing}, subspace clustering~\cite{you2016scalable,liu2013robust}, image deblurring~\cite{liu:tip:2014} and so on. To solve the problem in (\ref{Gene:L-ADMM}), the Linearized ADMM (LADMM) algorithm established by~\cite{lin2011linearized} is a desirable choice. But LADMM generally needs hundreds or more iterations to converge and is therefore time consuming; this motivates us to seek a learning-based version of LADMM. However, due to the presence of the equality constraint, existing theories are no longer applicable. As a consequence, we need to invent new algorithm design and theoretical analysis to address properly the following questions:
\begin{itemize}
\item[1.] How to combine the deep learning strategy with LADMM so as to solve the constrained problem in (\ref{Gene:L-ADMM})?
\item[2.] What is the relation between the learning-based LADMM and original LADMM? Specifically, does the output of learning-based LADMM still obey the linear constraint? And, most importantly, can the learning based LADMM still ensure convergence rate?
\end{itemize}
To make LADMM learnable, first of all, we convert the proximal operator in LADMM to a special neural network structure. Then we replace the given matrix $\*A$ and $\*B$ with some learnable weights and, meanwhile, expand the dimension of the penalty parameter such that the penalties on different directions are learnable as well, resulting in a novel method termed Differentiable LADMM (D-LADMM). What is more, we prove that, under some mild conditions, there do exist a set of learnable parameters that ensure D-LADMM to achieve a linear rate of convergence, and we show that those desired parameters can be attained by training D-LADMM properly. Interestingly, our results illustrate that it is possible for D-LADMM to possess a decay rate of linear convergence smaller than that of LADMM, which means that D-LADMM could converge faster than LADMM (note that LADMM is not linearly convergent unless the objective function is strongly convex). In summary, the main contributions of this paper include:
\begin{itemize}
	\item We propose a learning-based method called D-LADMM for solving the constrained optimization problem in (\ref{Gene:L-ADMM}). It is worth noting that our techniques, mainly including the proximal operator inspired network structure and the proposed policies for dealing with the linear constraint, would be useful for solving the other constrained problems.
	\item
	As for convergence, due to the high flexility of the learnable modules, it is difficult to assert the convergence of learning-based optimization. Remarkably, we establish a rigorous analysis on the convergence properties of D-LADMM. Our analysis shows that D-LADMM still satisfies the linear constraint and may converge faster than LADMM in some cases.
\end{itemize}
To the best of our knowledge, we are the \emph{first} to provide convergence analysis for learning-based optimization method under the context of constrained problems.

The remainder of the paper is organized as follows. We review some related work in Section~\ref{sec:Related}. In Section~\ref{sec:D-LADMM}, we start with a warm-up case to introduce how to convert the proximal operator in LADMM as a shallow neural network, and, accordingly, we establish the so-called D-LADMM. We analyze the convergence properties of D-LADMM in Section \ref{sec:Convergence}. Finally, empirical results that verify the proposed theories are given in Sections~\ref{sec:Experiments}.
\vspace{-2mm}
\section{Related Work}\label{sec:Related}
When ignoring the equality constraint of the problem in (\ref{Gene:L-ADMM}), there already exist some learning-based algorithms equal to the task, but most of them provide no convergence analysis. We have spotted only one theoretical article; namely,~\citet{chen2018theoretical} unroll the optimization procedure of LASSO as a RNN and prove that the resulted learning-based algorithm can achieve a linear rate of convergence. This result is significant but, to our knowledge, there is no similar conclusion available for constrained problems. The majority of the literature is consisting of empirical studies, e.g.,~\cite{ulyanov2018deep,zhang2017learning,diamond2017unrolled} consider DNN as implicit priors for image restoration. The problems addressed by these methods are in fact special cases of problem (\ref{Gene:L-ADMM}); namely, their formulations can be obtained by removing the constraint as well as some regularization term from (\ref{Gene:L-ADMM}). Due to the lack of theoretical analysis, it is unclear when or where their DNN dominant solution sequences should be terminated.
\begin{table*}[t]
	\centering
	\renewcommand\arraystretch{1.0}
	\caption{Summary of notations in this paper.}
	\label{tab:notations}
	\begin{tabular}{l|l|l|l}
		\hline
		$a$ & A scalar.  & $\mathbf{A}$ & A matrix.  \\
		$\mathbf{a}$  & A vector.  &$\#A(\cdot)$  & An operator.  \\ \hline
		$\*A\succ0$ & A positive-definite matrix.  & $\#A\succ0$ & A positive-definite operator, $\langle\#A(\*Z),\*Z\rangle>0,~\forall \*Z\neq 0$.  \\
		$\*I_m$ & $\*I_m\in \mathbb{R}^{m\times m}$ identity matrix.  &$\circ$  &Hadamard product~(entrywise product).  \\ \hline
		$\Vert \mathbf{a} \Vert_{2}$& $\Vert \mathbf{a} \Vert_{2}=\sqrt{\sum_{i}a_{i}^{2}}$.   &
		$\Vert \mathbf{A} \Vert_{F}$& $\Vert \mathbf{A} \Vert_{F}=\sqrt{\sum_{ij}A_{ij}^{2}}$ .  \\
		$\Vert \mathbf{A} \Vert_{1}$& $\Vert \mathbf{A} \Vert_{1}=\sum_{ij}\vert A_{ij}\vert$. &
		$\Vert \mathbf{A} \Vert$ & Maximum singular value.  \\
		$\|\bm{\omega}\|_{\#H}^2$ & $\left\langle\bm{\omega}, \#H(\bm{\omega})\right\rangle$.&
		$\operatorname{dist}_{\#H}^2(\bm{\omega}, \*\Omega^*)$& $\min_{\omega^* \in \*\Omega^*} \|\bm{\omega} - \bm{\omega}^*\|_{\#H}^2$ .  \\
		
		\hline

		$\#D$ & Linear operator defined in (\ref{Dk}).  &
		$\bm{\beta}$ & Positive matrix with ${\beta}_{ij}>0$.  \\
		$\bm{\omega}$ & $ \left(\*Z, \*E, -\bm{\lambda}\right)^\top$.  &
		$\*u$ & $ \left(\*Z, \*E\right)^\top$.  \\
		$\bm{\beta}^{-1}, 1/\bm{\beta}$ & The $ij$-th entry being $1/{\beta}_{ij}$.  &
		$\#H(\bm{\omega})$  & $\#H(\bm{\omega}) = \left(\#D(\*Z), \bm{\beta}\circ \*E, -(\bm{\beta})^{-1}\circ \bm{\lambda}\right)^\top$ . \\
		$h(\*u) $ & $f(\*Z) + g(\*E)$.  &
		$\#F_k(\bm{\omega})$ & $(\*W_k^\top\bm{\lambda}, \bm{\lambda}, \*{A}\*{Z}+\*B\*{E} - \*{X})^\top$. \\
		$\#G_k(\cdot)$ & $(\*W_k,\*I, \*0)^\top \bm{\beta}_k\circ(\cdot)$.&
		$\phi(\bm{\omega})$ & $(\*A^\top\bm{\lambda}+ \partial f(\*Z), \bm{\lambda} + \partial g(\*E), \*{A}\*{Z}+\*B\*{E} - \*{X})^\top$.\\
		$d^*(\cdot)$ & Lagrange dual function of (\ref{Gene:L-ADMM}).&
		$\*\Omega^*$ & The solution set of (\ref{Gene:L-ADMM}).
		\\ \hline
	\end{tabular}\vspace{-2mm}
\end{table*}

\citet{sun2016deep} recast the ADMM procedure as some learnable network, called ADMM-Net, and they apply it to a compressive sensing based Magnetic Resonance Imaging (MRI) problem that is indeed a special case of problem (\ref{Gene:L-ADMM}). The authors show that ADMM-Net performs well in MRI, but there is no guarantee for ensuring the convergence of their algorithm. Notice that our proposed D-LADMM is built upon LADMM rather than ADMM. Comparing to ADMM, LADMM needs fewer auxiliary variables to solve the constrained problems like (\ref{Gene:L-ADMM}). This detail is important, because fewer auxiliary variables means fewer learnable parameters while recasting the algorithm to DNN and, consequently, the reduction in the number of learnable parameters can accelerate the training process.
\vspace{-2mm}
\section{Differentiable Linearized ADMM}\label{sec:D-LADMM}
In this section, we shall begin with a warm-up case to show the conversion from a proximal iteration to some DNN block. Then we introduce the proposed D-LADMM in detail. The main notations used throughout this paper are summarized in Table~\ref{tab:notations}.
		
\subsection{Warm-Up: Differentiable Proximal Operator}
We first show how to differentialize the proximal operator as a network block. Consider the following unconstrained problem with two objective components:
\begin{equation}\label{PG}
\min_{\*z} f(\*z)+\frac{1}{2}\|\*A\*z -\*b\|_2^2,
\end{equation}
where $\*{A}\in \mathbb{R}^{m\times d}$, $\*{z} \in \mathbb{R}^{d}$, $\*{b} \in \mathbb{R}^{m}$, and $f(\cdot)$ is a real-valued convex function. The proximal gradient algorithm solves problem (2) as follows:
\begin{equation}\label{PGD}
\*z _ {  k  } = \textbf{prox} _ { t  f } \left( \*z _ {  k - 1  } - t\*A^\top (\*A \*z _ { k - 1} -\*b ) \right),
\end{equation}
where $t > 0$ is the step size and \textbf{prox} is the proximal operator given by
\begin{equation*}
\textbf{prox}_f(x) =  \argmin_{z}f(z) + \frac{1}{2}\|z-x\|_2^2.
\end{equation*}
As pointed out by~\cite{lin2011linearized,zhang2010bregmanized,blumensath2008iterative,bot2018proximal}, the \emph{strict} convergence of the proximal gradient procedure (\ref{PGD}) relies on the condition $t\|\*A\|^2<1$; that is,
\begin{equation}\label{PGD:Assumption}
\frac{1}{t}\*I - \*A^\top\*A\succ 0.
\end{equation}
The iteration (\ref{PGD}) is quite similar to a local block of DNN, i.e., $\*z _ {  k  } = \Phi(\*W_k\*z _{k-1} + \*b_k)$, where $\Phi(\cdot)$ is an NN block. Actually, the proximal operator connects tightly to the non-linear activation function in DNN. In some cases, they share the same form. For example, given $\*a \in \mathbb{R}^{d}$, we may set the function $f(\cdot)$ as $f(\*x) = \sum_{i=1}^d f_i(x_i)$, where
\[
f_i(x_i)  = \left\{ \begin{array} { l l } { \frac{1}{2}(x_i-a_i)^2 + c, } & { x_i<0, } \\ { 0, } & { x_i\geq 0. }\end{array} \right.
\]
Then the proximal operator coincides with ReLU, a commonly used non-linear activation function for DNN; namely, $\textbf{prox} _ { f}(\*a) = \operatorname{ReLU}(\*a)$, where $\operatorname{ReLU}(a_i) = \max\{a_i,0\}$. We can see that the proximal operator shares almost the same role as the activation function. Inspired by this, we can transform the iteration (\ref{PGD})  into a network structure, namely Differentiable Proximal Operator:
\begin{equation}\label{PGD3}
\*z_{k} = \zeta\left( \*z _ {  k - 1  } - \*W_{k-1}^\top (\*A \*z _ { k - 1} -\*b )\right),
\end{equation}
where $\zeta(\cdot)$ is some non-linear activation function and $\*W_{k-1}$ is the learnable parameter. As shown above, with proper $\zeta(\cdot)$ and function $f(\cdot)$, (\ref{PGD}) and (\ref{PGD3}) can be the same. 
\vspace{-2mm}\subsection{Differentiable Linearized ADMM}\vspace{-2mm}

\begin{figure*}[!t]
	\centering
	\vspace{-2mm}
	\includegraphics[width=0.85\linewidth]{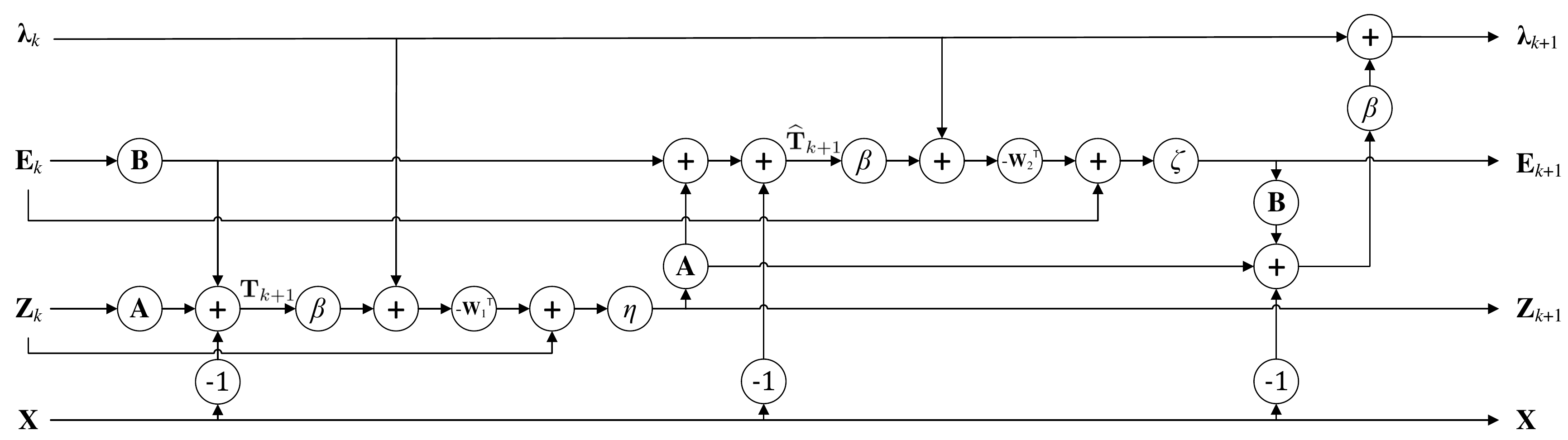}
	\vspace{-3mm}
	\caption{One block structure of the proposed D-LADMM. As we can see, such a LADMM inspired differentiable block reflects some prevalent structures, such as residual connection~\cite{he2016deep} and dense connection~\cite{huang2017densely}.}
	\label{fig:pipline}
	\vspace{-3mm}
\end{figure*}
First of all, we shall revisit the Linearized ADMM~(LADMM) algorithm~\cite{lin2011linearized}. Given $\*{A}\in \mathbb{R}^{m\times d_1}, \*B \in \mathbb{R}^{m\times d_2}$, $\*{X} \in \mathbb{R}^{m\times n}$ as well as two real-valued convex functions, $f(\cdot)$ and $g(\cdot)$, the iterative scheme of LADMM for solving (\ref{Gene:L-ADMM}) reads as:
\begin{equation}\label{L-ADMM_update}
\left\{
\begin{aligned}
&\*T_{k+1} = \*A\*Z_k + \*B\*E_k-\*X,\\
& \*{Z}_{k+1} = \textbf{prox}_{\frac{f}{L_1}}\left\{\*{Z}_k - \frac{1}{L_1}\*A^\top(\bm{\lambda}_k + \beta \*T_{k+1} )\right\},\\
&\widehat{\*T}_{k+1} = \*A\*Z_{k+1} + \*B\*E_k-\*X,\\
& \*E_{k+1} = \textbf{prox}_{\frac{g}{L_2}}\left\{\*{E}_k - \frac{1}{L_2}\*B^\top(\bm{\lambda}_k + \beta \widehat{\*T}_{k+1} )\right\},\\
& \bm{\lambda}_{k+1} = \bm{\lambda}_k + \beta (\*A\*Z_{k+1} + \*B\*E_{k+1}-\*X),
\end{aligned}
\right.
\end{equation}
where $\bm{\lambda}$ is Lagrange multiplier, $L_1>0$ and $L_2>0$ are Lipsitz constants, and $\beta >0$ is penalty parameter.

Following the spirits of learning-based optimization, we propose a DNN named Differentiable LADMM (D-LADMM). One iteration of the original LADMM is generalized as a block of neural network. Specifically, we retain the updating rule for $\*T_{k+1}$, $\widehat{\*T}_{k+1}$ and the Lagrange multiplier $\bm{\lambda}$, replace the two proximal steps in (\ref{L-ADMM_update}) by differentiable proximal operators and, meanwhile, expand the dimension of the penalty parameter $\beta$ such that the penalties on different directions are learnable as well. In summary, one block of our D-LADMM is given by 
\begin{equation}
\left\{
\begin{aligned}\label{D-LADMM_update}
& \*T_{k+1} = \*A\*Z_k + \*B\*E_k-\*X,\\
& \*{Z}_{k+1} = \eta_{(\bm{\theta}_1)_k}\left(\*{Z}_k - (\*W_1)_k^\top(\bm{\lambda}_k + \bm{\beta}_k \circ \*T_{k+1}) \right),\\
&\widehat{\*T}_{k+1} = \*A\*Z_{k+1} + \*B\*E_k-\*X,\\
& \*{E}_{k+1} = \zeta_{(\bm{\theta}_2)_k}\left(\*{E}_k - (\*W_2)_k^\top(\bm{\lambda}_k + \bm{\beta}_k \circ \widehat{\*T}_{k+1}) \right),\\
& \bm{\lambda}_{k+1} = \bm{\lambda}_k + \bm{\beta}_k \circ (\*A\*Z_{k+1} + \*B\*E_{k+1}-\*X),
\end{aligned}
\right.
\end{equation}
where $\Theta = \{(\*W_1)_k, (\*W_2)_k, (\bm{\theta}_1)_k, (\bm{\theta}_2)_k, \bm{\beta}_k\}_{k=0}^{K}$ are learnable matrices, and $\circ$ is the element-wise product. In addition, $\eta(\cdot)$ and $\zeta(\cdot)$ are some non-linear functions parameterized by $\bm{\theta}_1$ and $\bm{\theta}_2$, respectively.
\par
It is worth mentioning that we intentionally keep the matrices $\*A$ and $\*B$ in the updating step for $\*T_{k+1}$, $\widehat{\*T}_{k+1}$ and $\bm{\lambda}$. The reason is that, instead of leaving everything for NN to learn, our scheme can in fact benefit the compliance of the equality constraint in (\ref{Gene:L-ADMM}) so as to help reveal the connection between D-LADMM and LADMM.
\par
D-LADMM, a block of which is illustrated in Figure~\ref{fig:pipline}, actually corresponds to a $K$-layer feed-forward neural network with side connections. Many empirical results, e.g., \cite{gregor2010learning,wang2016d3,sun2016deep,peng2018k}, show that a well-trained $K$-layer differentiable optimization inspired model---compared with the original optimization algorithm---can obtain almost the same good solution within one or even two order-of-magnitude fewer iterations. Moreover, the quality of the output from each layer will being gradually improved.
\par
Among the other things, it is also feasible to expand the parametric space of D-LADMM by introducing more learnable units. For example, we can generalize the linear transformation $\*{A}$ to some non-linear mapping $\mathcal{A}_{\bm{\vartheta_1}}(\cdot): \mathbb{R}^{d} \rightarrow \mathbb{R}^{m}$ parameterized by $\bm{\vartheta_1}$, so as to learn adaptively some proper representation from data. But a theoretical analysis for such models has to be much more involved.
\par
\textbf{Training Strategy:} Different from LADMM which has no parameter to learn, D-LADMM is treated as a special structured neural network and trained using stochastic gradient descent (SGD) over the observation $\*X$, and all the parameters $\Theta$ are subject to learning. Depending on whether the ground truth (i.e., true optimal solution to (\ref{Gene:L-ADMM})) is given, we present two different training strategies. Without the ground truth, the training process is actualized by 
\vspace{-1mm}
\begin{equation}\label{D-LADMM_loss1}
\min_{\Theta}  f(\*Z_K) + g(\*E_K) - d^*(\bm{\lambda}_{K}),
\end{equation}
where $d^*(\bm{\lambda}_{K})$ is the dual function of (\ref{Gene:L-ADMM}) defined as $d^*(\bm{\lambda}_{K}) = \inf_{\*Z,\*E} f(\*Z) + g(\*E) + \left\langle  \bm{\lambda}_{K}, \*A\*Z + \*B\*E-\*X\right\rangle$. When the functions $f(\cdot)$ and $g(\cdot)$ are given, we can obtain explicitly the dual function $d^*(\cdot)$, which is a concave function bounded from above by $f(\cdot)+ g(\cdot)$; this means the objective in (\ref{D-LADMM_loss1}) is always non-negative. Moreover, due to the convexity of $f(\cdot)+ g(\cdot)$, the minimum attainable value of (\ref{D-LADMM_loss1}) is exactly zero. In other words, the global optimum is attained whenever the objective reaches zero.
\par
In the cases where the ground-truth $\mathbf{Z}^{*}$ and $\mathbf{E}^{*}$ are provided along with the training samples, D-LADMM can be trained by simply minimizing the following square loss:
\begin{equation}\label{D-LADMM_loss2}
\min_{\Theta}  \|\*Z_K - \*Z^*\|_F^2 + \|\*E_K - \*E^*\|_F^2.
\end{equation}

\vspace{-5mm}
\section{Convergence Analysis}
\label{sec:Convergence}
In this section, we formally establish the convergence of the proposed D-LADMM, based on some specific settings. Although learning-based optimization methods have achieved a great success in practice, their merits have not been validated sufficiently from the theoretical perspective.
\citet{chen2018theoretical} and~\citet{liu2018alista} 
provide the convergence analysis for Unfolded Iterative Shrinkage and Thresholding Algorithm (ISTA). However, their techniques are applicable only to unconstrained problems and thus not very helpful for analyzing D-LADMM, which contains an equality constraint and some complex updating rules. In general, due to the presence of non-linear learnable functions, it is hard, if not impossible, to analyze theoretically the model in (\ref{D-LADMM_update}). Fortunately, with some proper settings, we can still accomplish a rigorous analysis for D-LADMM.

\subsection{Settings}
Please notice that the analytical approach remains the same while one of $\*A$ and $\*B$ is suppressed as the identity matrix. Thus, for the sake of simplicity, we omit the given matrix $\*B$ from problem (\ref{Gene:L-ADMM}). We shall also focus on the cases where $\eta$ and $\zeta$ are respectively the proximal operators of the functions $f(\cdot)$ and $g(\cdot)$. In other words, we consider the following simplified D-LADMM for analysis:
\begin{equation}
\left\{
\begin{aligned}\label{D-LADMMCov}
& \*T_{k+1} = \*A\*Z_k + \*E_k-\*X,\\
& \*{Z}_{k+1} = \textbf{prox}_{\!f_{\bm{\theta}_k}\!}\left(\*{Z}_k \!-\! \frac{1}{\bm{\theta}_k}\!\circ\![\*W_k^\top(\bm{\lambda}_k + \bm{\beta}_k \circ \*T_{\!k+1\!})] \right),\\
& \*E_{k+1} = \textbf{prox}_{g_{\bm{\beta}_k}}\left(\*X-\*A\*Z_{k+1} - \frac{1}{\bm{\beta}_k}\circ\bm{\lambda}_k \right),\\
& \bm{\lambda}_{k+1} = \bm{\lambda}_k + \bm{\beta}_k \circ (\*A\*Z_{k+1} + \*E_{k+1}-\*X),
\end{aligned}
\right.
\end{equation}
where $\textbf{prox}_{g_{\bm{\beta}}}(\*R) = \argmin_{\*E}\{g(\*E) + \frac{\bm{\beta}}{2}\circ \|\*E-\*R\|_F^2\}$ and $\bm{\beta}_k, \bm{\theta}_k>0$.

By the definition of the proximal operator, we rewrite the above (\ref{D-LADMMCov}) as follows:
\begin{equation}
\left\{
\begin{aligned}\label{D-LADMMCov2}
&\*{Z}_{k+1} = \argmin_{\*Z}\left\{f(\*Z)  + \frac{\bm{\theta}_k}{2}\circ\|\*Z - \*{Z}_k + (\bm{\theta}_k)^{-1}\circ \right.\\
&\qquad\qquad~ \left. \*W_k^\top \big(\bm{\lambda}_k + \bm{\beta}_k\circ( \*A\*Z_k + \*E_k-\*X)\big)\|_F^2 \right\},\\
& \*E_{k+1} = \argmin_{\*E}\left\{g(\*E) + \frac{\bm{\beta}_k}{2}\circ\|\*E-\*X+\*A\*Z_{k+1} \right.\\
& \qquad\qquad\qquad\qquad\qquad\qquad\qquad \left. + ({\bm{\beta}_k})^{-1} \circ \bm{\lambda}_k\|_F^2 \right\},\\
& \bm{\lambda}_{k+1} = \bm{\lambda}_k + \bm{\beta}_k\circ (\*A\*Z_{k+1} + \*E_{k+1}-\*X).
\end{aligned}
\right.
\end{equation}
The above is indeed a special case of D-LADMM (\ref{D-LADMM_update}), with learnable parameters $\Theta = \{(\*W_k\in \mathbb{R}^{m\times d}, \bm{\theta}_k\in \mathbb{R}^{d\times n}, \bm{\beta}_k\in \mathbb{R}^{m\times n})\}_{k=0}^{K}$.
\par
As aforementioned, the success of proximal operator based methods may rely on the condition in (\ref{PGD:Assumption}). Given $(\*W, \bm{\theta}, \bm{\beta})$, the positive-definite matrix in (\ref{PGD:Assumption}) becomes a linear operator $\#D: \mathbb{R}^{d\times n} \rightarrow \mathbb{R}^{d\times n}$ given by 
\begin{equation}\label{Dk}
\#D(\*Z) = \bm{\theta} \circ(\*Z) - \*W^\top\bm{\beta}\circ(\*A\*Z), ~\forall \*Z \in \mathbb{R}^{d\times n}.
\end{equation}
It is known that, to ensure the convergence of LADMM, the positive-definiteness condition in (\ref{PGD:Assumption}) has to be obeyed. For the same reason, the convergence of D-LADMM necessitates the positive-definiteness of the operator $\#D$. Based on this, we define the following set:
\begin{equation}
\begin{aligned}\label{fea_set}
\mathcal{S}(\sigma, \*A) \triangleq \left\{(\*W, \bm{\theta}, \bm{\beta})
\big|\right.&\left.\|\*W-\*A\|\leq\sigma, \#D\succ0,\right.\\
&~~\left.\bm{\beta}, \bm{\theta}>0\right\},
\end{aligned}
\end{equation}
where $\#D\succ0$ means that the operator $\#D$ is a positive-definite operator. The above set, $\mathcal{S}(\sigma, \*A)$, could be non-empty if a proper $\*A$ is given. For example, when $\*W = \*A$ and both $\bm{\beta}$ and $\bm{\theta}$ degenerate to scalars, the non-emptiness of $\mathcal{S}(\sigma, \*A)$ is actually equivalent to the classical condition (\ref{PGD:Assumption}). In general, $\|\*W-\*A\|<\sigma$ ensures that the learnable weight is close to $\*A$, which guarantees that the first minimization problem in (\ref{D-LADMMCov2}) has an analogous optimization landscape with the original one, and $\#D\succ0$ ensures that the local approximation in the proximal operator is a strict upper bound of the original objective. Thus, we can treat the non-emptiness of the set $\mathcal{S}(\sigma, \*A)$ as a generalization of the condition (\ref{PGD:Assumption}), and we would make an assumption as follows.
\begin{assumption}\label{asum2}{
There exists a constant $c$ such that $\mathcal{S}(\sigma, \*A)$ is non-empty for any $\sigma$ that satisfies $0\leq \sigma \leq c$, namely the given $\*A$ is proper.
}\end{assumption}
\vspace{-3mm}\subsection{Convergence Property}\vspace{-1mm}
Before proving the main theorem, we shall establish some basic lemmas by which the candidate parameters set for $\Theta = \{\*W_k, \bm{\theta}_k, \bm{\beta}_k\}_{k=0}^{K}$ can be derived. Note that, for all the lemmas and theorems in this section, we assume that the Assumption \ref{asum2} is satisfied.
\par
Define an operator $\#H_k$ as 
\begin{equation}\label{H_def1}
 \#H_k(\cdot) =
  \left(
  \begin{array}{ccc}
          \#D_k & 0& 0 \\
          0& \bm{\beta}_k\circ& 0\\
          0& 0& (\bm{\beta}_k)^{-1}\circ
 \end{array}
 \right)(\cdot),
\end{equation}
where $\#D_k$ is the operator (\ref{Dk}) given $(\*W_k, \bm{\theta}_k, \bm{\beta}_k)$.

Firstly, we prove an inequality for the objective $h(\*u_k)$.
\begin{lem}\label{convexCond}
Let the sequence $\{\bm{\omega}_k\}$ be generated by (\ref{D-LADMMCov2}). Then we have:
\begin{equation}\label{bound1}
\begin{aligned}
h(\*u) &- h(\*u_{k+1}) + \left\langle\bm{\omega}-\bm{\omega}_{k+1},\#F_k(\bm{\omega}_{k+1})+\right.\\
&\left.\#G_k(\*E_k-\*E_{k+1})+ \#H_k(\bm{\omega}_{k+1} - \bm{\omega}_k)\right\rangle \geq 0, ~~\forall \bm{\omega},
\end{aligned}
\end{equation}
where $\#F_k(\cdot)$ and $\#G_k(\cdot)$ are simply two linear operators as defined in Table \ref{tab:notations}.
\end{lem}

Lemma \ref{convexCond} suggests that the quantity $\|\bm{\omega}_{k+1} - \bm{\omega}_k\|_{\#H_k}^2$ could be used to measure
the distance between the iterate $\bm{\omega}_{k+1}$ to the solution set $\*\Omega^*$. In other words, when $\|\bm{\omega}_{k+1} - \bm{\omega}_k\|_{\#H_k}^2 = 0$, the positive-definiteness of $\#H_k(\cdot)$ gives $\bm{\omega}_{k+1} - \bm{\omega}_k = 0$. If $\*W_{k} = \*A$, we have
\[
h(\*u)- h(\*u_{k+1}) + \left\langle\bm{\omega}-\bm{\omega}_{k+1},\#F_k(\bm{\omega}_{k+1})\right\rangle \geq 0, ~~\forall \bm{\omega},
\]
which implies that $\bm{\omega}_{k+1}$ is a solution of problem (\ref{Gene:L-ADMM}).

\begin{lem}\label{Decent}
Let the sequence $\{\bm{\omega}_k\}$ be generated by (\ref{D-LADMMCov2}). Suppose that, for any point $\bm{\omega}^*\in \*\Omega^*$, there exists proper $(\*W_k, \bm{\theta}_k, \bm{\beta}_k) \in \mathcal{S}(\sigma, \*A)$ such that:
\begin{equation}
\left\langle\bm{\omega}_{k+1} - \bm{\omega}^*, \#H_k(\bm{\omega}_k - \bm{\omega}_{k+1})\right\rangle\geq 0, ~\forall k\geq0,
\label{bound2}
\end{equation}
where $\#H_k(\cdot)$ is given in (\ref{H_def1}). Then $\|\bm{\omega}_k\|_F<\infty$ holds for all $k$, and we have:
\begin{equation}\label{bound3}
\|\bm{\omega}_k - \bm{\omega}^*\|_{\#H_k}^2 \geq \|\bm{\omega}_{k+1} - \bm{\omega}^*\|_{\#H_k}^2 + \|\bm{\omega}_k - \bm{\omega}_{k+1}\|_{\#H_k}^2.
\end{equation}
\end{lem}
Lemma~\ref{Decent} shows that there exist proper learnable parameters that make $\{\bm{\omega}_k\}$ strictly contractive with respect to the solution set $\*\Omega^*$. It is worth mentioning that the proof of Lemma \ref{Decent} may partly explain why D-LADMM converges faster than LADMM. Denote $\|\*W_k - \*A\| = \sigma_k$. From the proof process, we find that, when $\|\*E_{k+1} -\*E_{k}\|_F$ is large and $\|\*Z_{k+1} - \*Z^*\|_F$ is small, $\sigma_k$ can be set as a large value, which means the feasible space of the learnable weight is large as well. Conversely, the better weight retrieved from the larger feasible space can also promote the convergence speed. In one word, the sequence $\{\sigma_k\}_{k=0}^K$ is somehow learnt adaptably so as to benefit global convergence.\par
We now show that, whenever there is no solution that satisfies $(\*W_k, \bm{\theta}_k, \bm{\beta}_k) \in \mathcal{S}(\sigma, \*A)$ and $\bm{\omega}_{k+1}\neq\bm{\omega}_k$, the optimum is attained.
\begin{lem}\label{non-zero}
Let the sequence $\{\bm{\omega}_k\}$ be generated by (\ref{D-LADMMCov2}). Given $\bm{\omega}_k$, if the updating rule in (\ref{D-LADMMCov2}) achieves $\bm{\omega}_{k+1} = \bm{\omega}_k$ for all $(\*W_k, \bm{\theta}_k, \bm{\beta}_k) \in \mathcal{S}(\sigma, \*A)$, then $\bm{\omega}_k = \bm{\omega}_{k+1} \in \*\Omega^*$.
\end{lem}
\vspace{-3mm}\subsection{Main Results}\vspace{-1mm}
Equipped with the above lemmas, we establish a couple of theorems to guarantee the convergence of D-LADMM (\ref{D-LADMMCov2}).
\begin{theorem}[Convergence of D-LADMM]\label{main_thm1}{
Let the sequence $\{\bm{\omega}_k\}$ be generated by (\ref{D-LADMMCov2}). There exists proper $(\*W_k, \bm{\theta}_k, \bm{\beta}_k) \in \mathcal{S}(\sigma, \*A)$ such that $\{\bm{\omega}_k\}$ converges to a solution $\bm{\omega}^*\in\*\Omega^*$.
}\end{theorem}\vspace{-2mm}
So, in general, there exist proper parameters such that the outputs of our D-LADMM converge to the optimal solution of problem (\ref{Gene:L-ADMM}). In the rest of this subsection, we will further investigate its convergence rate, which is measured by a point-to-set distance $\operatorname{dist}_{\#H}^2(\bm{\omega}, \*\Omega^*)$. The following theorem shows that we can find a set of parameters to make the distance decrease monotonically.
\begin{theorem}[Monotonicity of D-LADMM]\label{main_thm2}{
Let the sequence $\{\bm{\omega}_k\}$ be generated by (\ref{D-LADMMCov2}). There exists proper $(\*W_k, \bm{\theta}_k, \bm{\beta}_k) \in \mathcal{S}(\sigma, \*A)$ such that $\operatorname{dist}_{\#H_k}^2(\bm{\omega}_k, \*\Omega^*)$ decreases monotonically when $k$ is large enough.
}\end{theorem}\vspace{-2mm}
The above theorem proves the monotonic decreasing property of the deviation between the produced solution $\bm{\omega}_k$ and the true solution set $\*\Omega^*$. This is different from Lemma \ref{Decent} in which the distance is measured by a point-to-point metric. For convenience, we combine all the updating rules in (\ref{D-LADMMCov2}) into a single operator $\#T$:
$$\mathcal{T}(\*W_k, \bm{\theta}_k, \bm{\beta}_k)(\bm{\omega}_k) = \bm{\omega}_{k+1}.$$
Next, we will show that, under some condition imposed on $\#T$, D-LADMM can attain a linear rate of convergence.
\begin{theorem}[Convergence Rate of D-LADMM]\label{main_thm3}{
Let the sequence $\{\bm{\omega}_k\}$ be generated by (\ref{D-LADMMCov2}). Suppose that there exist $(\*A, \bm{\theta}^*, \bm{\beta}^*)$ and $K_0>0$ such that for any $k\geq K_0$ (i.e., $k$ is large enough) the following holds:
\begin{equation}\label{ErrBound}
\operatorname{(EBC):\qquad}\operatorname{dist}_{\#H^*}^2(\widetilde{\bm{\omega}}, \*\Omega^*)\leq \frac{\kappa}{16} \|\widetilde{\bm{\omega}} - \bm{\omega}_k\|_{\#H^*}^2,
\end{equation}
where $\#H^*(\cdot)$ is given in (\ref{H_def1}) by setting $(\*W_k, \bm{\theta}_k, \bm{\beta}_k)$ as $(\*A, \bm{\theta}^*, \bm{\beta}^*)$ and $\widetilde{\bm{\omega}} = \#T(\*A, \bm{\theta}^*, \bm{\beta}^*)(\bm{\omega}_k)$. Then there exists proper $(\*W_k, \bm{\theta}_k, \bm{\beta}_k) \in \mathcal{S}(\sigma, \*A)$ such that $\operatorname{dist}_{\#H_k}^2(\bm{\omega}, \*\Omega^*)$ converges to zero linearly; namely, 
\begin{equation}\label{LinearCov}
\operatorname{dist}_{\#H_{k+1}}^2(\bm{\omega}_{k+1}, \*\Omega^*)< \gamma~ \operatorname{dist}_{\#H_k}^2(\bm{\omega}_k, \*\Omega^*),
\end{equation}
where $\gamma$ is some positive constant smaller than 1.
}\end{theorem}\vspace{-2mm}
From Theorem \ref{main_thm3}, we know that if the Error Bound Condition (EBC) in (\ref{ErrBound}) is satisfied, then there exists a sequence $\{(\*W_k, \bm{\theta}_k, \bm{\beta}_k)\}_{k=0}^{K} \rightarrow (\*A, \bm{\theta}^*, \bm{\beta}^*)$ that entables the linear decay rate of $\operatorname{dist}_{\#H_k}^2(\bm{\omega}_k, \*\Omega^*)$. Actually, our learning based D-LADMM does obtain a faster convergence speed than fixing all parameters as $(\*A, \bm{\theta}^*, \bm{\beta}^*)$. To confirm this, we give a specific example in the following lemma.
\begin{lem}\label{D-LADMM:example}
Consider the case where the \textbf{prox} operator of the function $f(\cdot)$ is bijective. For any $\bm{\omega}\not\in\*\Omega^*$ and $\bm{\omega}^* \in \*\Omega^*$, there exists a residue $(\*\Delta_w, \*\Delta_\theta, \*\Delta_\beta)$ such that $\|\#T(\*A+\*\Delta_w, \bm{\theta}^*+\*\Delta_\theta, \bm{\beta}^*+\*\Delta_\beta)(\bm{\omega}) - \bm{\omega}^*\|_F<\|\#T(\*A, \bm{\theta}^*, \bm{\beta}^*)(\bm{\omega}) - \bm{\omega}^*\|_F$.
\end{lem}\vspace{-2mm}
Lemma \ref{D-LADMM:example} implies that, at each iteration, we can find appropriate parameter to construct a solution that is closer to $\*\Omega^*$ than the solution produced by fixed parameters. Hence, it is entirely possible for the proposed D-LADMM to achieve a convergence rate higher than the traditional LADMM. The first experiment in Section \ref{sec:Experiments} verifies the superiority of D-LADMM over LADMM, in terms of convergence speed.
\par
It is worth noting that, although the convergence speed of LADMM can be improved by setting suitable penalty parameter $\beta$ for each dimension, it is really difficult to do this in practice. In contrast, the proposed D-LADMM provides a convenient way to learn all parameters adaptively from data so as to gain high convergence speed, as will be confirmed in the experiments.
\vspace{-2mm}\subsection{Discussions on D-LADMM}\vspace{-1mm}
First, our analysis for the convergence of D-LADMM is very general and can actually include the traditional LADMM as a special case. In effect, some already known properties of LADMM can be deduced from our results, but not vice versa. The analysis techniques designed for traditional optimization cannot be directly applied to our case, and it is considerably more
challenging to establish the convergence properties of D-LADMM.
\par
\begin{table*}[!tp]
	\small
	\renewcommand\arraystretch{1.1}
	\caption{PSNR comparison on $12$ images with noise rate $10\%$. For LADMM, we examine its performance at a couple of different iterations. LADMM is comparable to D-LADMM only when it undergoes a large number of iterations.}
	\setlength{\tabcolsep}{4pt}
	\begin{tabular}{l|c|c|c|c|c|c|c|c|c|c|c|c|c|}
		\cline{2-14}
		& \multirow{2}{*}{PSNR} & \multicolumn{12}{c|}{Images}                                                                              \\ \cline{3-14}
		&                       & Barb & Boat & France & Frog & Goldhill & Lena & Library & Mandrill & Mountain & Peppers & Washsat & Zelda \\ \cline{2-14}
		& Baseline              & 15.4 & 15.3 &  14.5  & 15.6 &  15.4    & 15.4 &  14.2   &  15.6    & 14.4     & 15.1    &  15.1   &  15.2 \\ \cline{2-14}
		& LADMM~(iter=15)                 & 22.1 & 24.2 &  18.0  & 23.1 &  25.2    & 25.6     &  15.0   &  21.7    & 17.7     & 25.1    &  30.6   &  29.7 \\ \cline{2-14}
		& LADMM~(iter=150)                 & 27.9 & 29.8 &  21.6  & 26.5 &  30.4    & 31.3 &  17.8   &  24.3    & 20.5     & 30.0    &  34.5   &  35.7 \\ \cline{2-14}
		& LADMM~(iter=1500)                 & 29.9 & 31.1 &  22.2  & 26.9 &  31.8    & 33.2 &  18.0   &  25.1    & 20.7     & 32.8    &  36.2   &  37.8 \\ \cline{2-14}
		& D-LADMM~($K$=15)               &  29.5 & 31.3 &  21.9      &  25.9    &  32.5        & 35.1 &  18.8       &  24.5        &    19.3      &    34.3     &  35.6  & 38.9      \\ \cline{2-14}
	\end{tabular}
	\label{tab:PSNR}
	\vspace{-3mm}
\end{table*}
Second, our EBC is indeed weaker than the conditions assumed by the previous studies (e.g.,~\cite{han2013local,han2015linear,yang2016linear}), which prove linear convergence rate for ADMM or LADMM. More precisely, as pointed out by~\citet{liu2018partial}, all the EBCs in~\cite{han2013local,han2015linear,yang2016linear} can be equivalently expressed as
\begin{equation}\label{ErrBoundOld}
\operatorname{dist}(\bm{\omega}, \*\Omega^*)\leq \kappa \cdot \|\phi(\bm{\omega})\|_F, ~\forall \bm{\omega},~\operatorname{dist}(\bm{\omega}, \*\Omega^*)\leq \epsilon,
\end{equation}
where $\phi(\cdot)$ is a mapping given in Table \ref{tab:notations}. Notably, the following lemma confirms the generality of our EBC. 
\begin{lem}\label{lem:weakercond}
The EBC in (\ref{ErrBoundOld}) suffices to ensure the validity of our EBC given in (\ref{ErrBound}).
\end{lem}\vspace{-2mm}
One may have noticed that our EBC is somewhat similar to the condition by~\citet{liu2018partial}. Yet the work of~\cite{liu2018partial} did not reveal the merit of learning-based optimization; that is, as shown in Lemma~\ref{Decent}, all the parameters are learnt automatically from data with the purpose of accelerating the convergence speed.
\begin{figure}[!tp]
	\centering
	\includegraphics[width=0.75\linewidth]{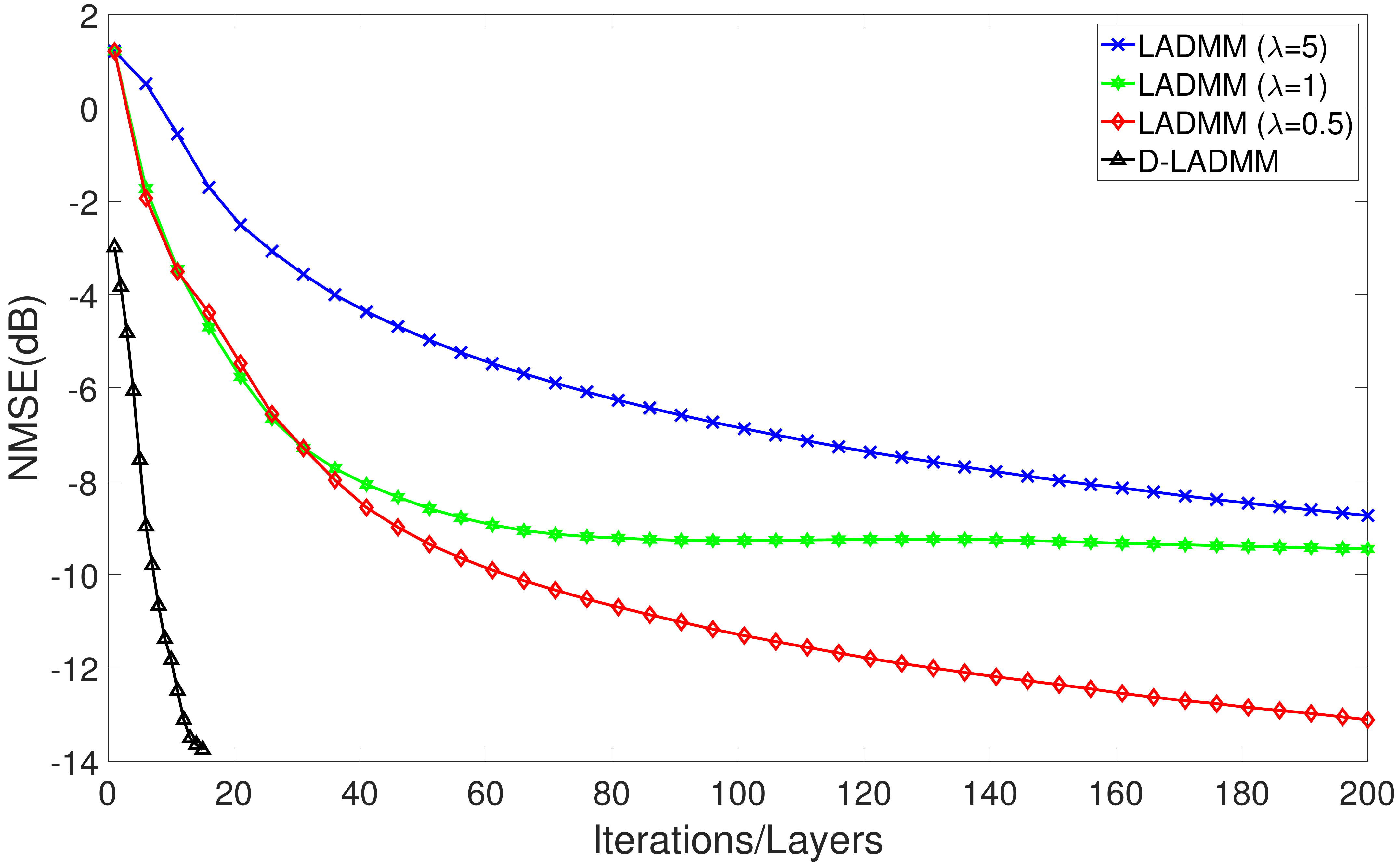}\\
	\vspace{-4mm}
	\caption{NMSE comparison among D-LADMM and LADMM with different $\lambda$ on the simulation dataset.}\label{fig:Simulate_loss}
	\vspace{-4mm}
\end{figure}
\vspace{-2mm}
\section{Experiments}\vspace{-1mm}
\label{sec:Experiments}
\begin{figure*}[t]
	\centering
	\subfigure[GND]{
		{\label{fig:GND_Lena}} 
		\includegraphics[width=1.52in]{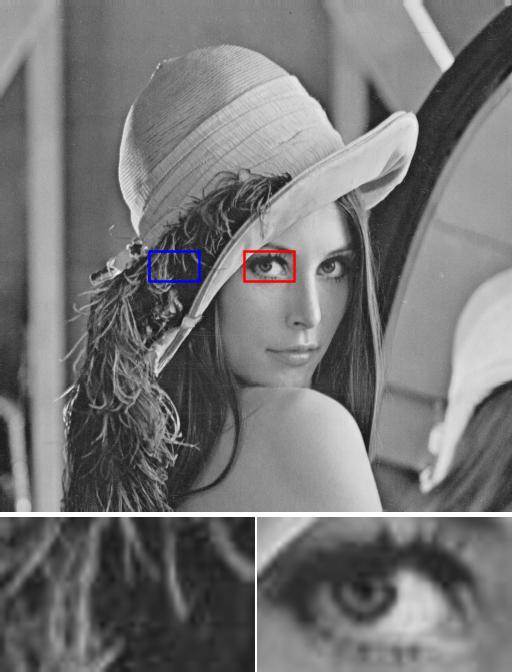}}
	\hspace{-1mm}
	\subfigure[Noisy image~(PSNR=15.4)]{
		{\label{fig:Noisy_Lena}} 
		\includegraphics[width=1.52in]{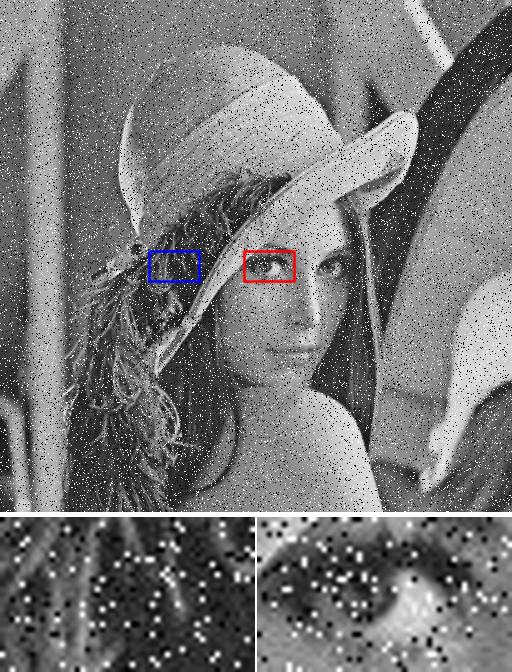}}
	\hspace{-1mm}
	\subfigure[D-LADMM~(PSNR=35.1)]{
		{\label{fig:D-LADMM_Lena}} 
		\includegraphics[width=1.52in]{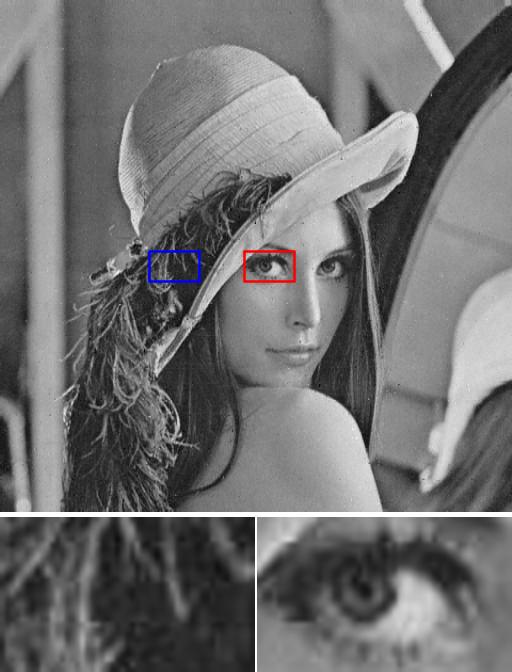}}
	\hspace{-1mm}
	\subfigure[LADMM~(PSNR=31.3)]{
		\label{fig:LADMM_Lena} 
		\includegraphics[width=1.52in]{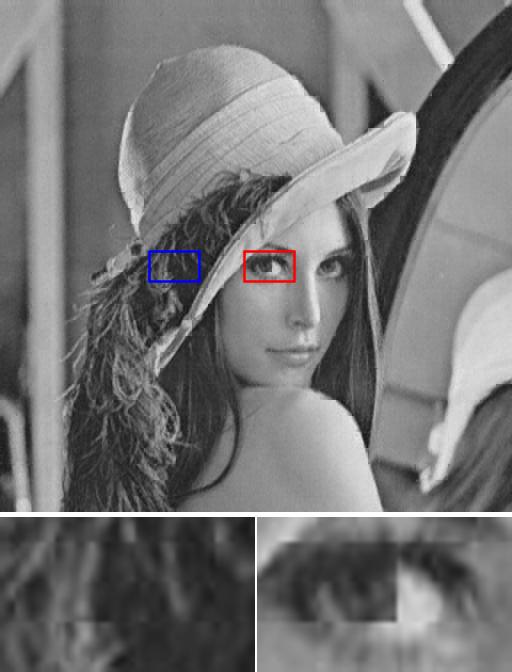}}
	\vspace{-3mm}
	\caption{Denoising results the Lena image. (a)  The ground-truth images. (b) The noisy images with salt-and-pepper noise of rate $10\%$. (c) Denoised images by our D-LADMM. (d) Denoised images by LADMM with 150 iterations. Best view on screen!}
	\label{fig:MONARCH} 
	\vspace{-3mm}
\end{figure*}
Different from LADMM which contains no learnable parameter, the proposed D-LADMM is treated as a structured neural network and trained using stochastic gradient descent (SGD) over the observation $\*X$. All the parameters, denoted as $\Theta$, are subject to learning. For convenience, we mainly consider the following $\ell_{1}$-norm constrained problem for empirical validations:
\begin{equation}\label{SSC}
	\min_{\*{Z}, \*{E}}~  \lambda \Vert \*{Z} \Vert_{1} +  \Vert \*{E}\Vert_{1}, ~~~\ s.t. \ \*{X}=\*{A}\*{Z}+\*{E},
\end{equation}
where $\lambda$ is a parameter to balance the contribution of each term. We use both LADMM and the proposed D-LADMM to solve the above problem, and compare their results on both synthetic datasets and natural images\footnote{Code:  {\color{blue}https://github.com/zzs1994/D-LADMM}}. Our D-LADMM is implemented on the PyTorch platform.
\vspace{-1mm}
\subsection{Simulation Experiments}
We first experiment with synthetic data, using similar experimental settings as~\cite{chen2018theoretical}. Specifically, we set $m=500$ and $d=250$. The numbers of training and testing samples are set to $10,000$ and $1,000$, respectively. Elements of the dictionary matrix $\mathbf{A}$ are sampled from i.i.d. Gaussian, namely $A_{i,j}\sim N(0, 1/d)$. The columns of $\mathbf{A}$ are normalized have a unit $\ell_{2}$ norm. To make a fair comparison, $\mathbf{A}$ is fixed and shared by all considered methods. The sparse coefficient matrix $\mathbf{Z}$ is generated by using a Bernoulli sampling operator (with probability 0.1) to randomly select values from the standard Gaussian distribution, i.e., $\mathbf{Z} = \mathrm{Ber}(0.1)\circ{}N(0,1)$. The sparse matrix $\mathbf{E}$ is generated in the same way as $\mathbf{Z}$, and the data samples for training and testing are constructed as $\mathbf{X} = \mathbf{AZ} + \mathbf{E}$.
\par
For the proposed D-LADMM, the number of layers is set to $K=15$. SGD is adopted to update the parameters with learning rate ${lr} = 0.01$. Regarding the activation function, we use the softshrink operator by~\citet{beck2009fast}. In these experiments, the ground-truth $\mathbf{Z}^{*}$ and $\mathbf{E}^{*}$ of training samples are known, thereby the second strategy in~(\ref{D-LADMM_loss2}) is adopted to train our D-LADMM network. The results are evaluated by a measure of NMSE~(normalized mean square error in dB), defined in terms of both $\mathbf{Z}$ and $\mathbf{E}$:
\begin{equation}
	\text{NMSE} \! = \!   10 \log_{10} \! \left( \frac{ \Vert \mathbf{Z}_K  \! - \!  \mathbf{Z}^{*} \Vert_{F}^{2}}{ \Vert \mathbf{Z}^{*} \Vert_{F}^{2}} \!  + \! \frac{\Vert \mathbf{E}_K \!  - \!  \mathbf{E}^{*} \Vert_{F}^{2}}{ \Vert \mathbf{E}^{*} \Vert_{F}^{2}} \right)\! .
	\label{NMSE}
\end{equation}

In Figure~\ref{fig:Simulate_loss}, we compare the proposed D-LADMM with LADMM. As one can see, the NMSE achieved by D-LADMM decreases linearly as the layer number $k$ grows and is much smaller than that by LADMM. These results confirm our main results stated in Theorems~\ref{main_thm1},~\ref{main_thm2} and~\ref{main_thm3}. Note here that, unlike D-LADMM, LADMM needs a proper $\lambda$ to produce correct solutions, thereby we test several different $\lambda$'s for LADMM. One can see that the choice of $\lambda$ has dramatic influences on the performance of LADMM, and smaller $\lambda$ may lead to better results.
\vspace{-1mm}
\subsection{Natural Image Denoising}
We also evaluate the considered methods on the task of natural image denoising, which is to remove the noise term $\mathbf{E}$ from the noisy observation $\mathbf{X}$, or recover the noise-free image $\mathbf{AZ}$ from $\mathbf{X}$ as equal. The experimental data is a classic dataset consisting of 12 natural images, called Waterloo BragZone Greyscale set2, in which a fraction of $r\%$ Salt-and-pepper noise is added to each image. Furthermore, the rectangle of each image is divided into non-overlapping patches of size $16 \times 16$. We use the patch-dictionary method~\cite{xu2014fast} to learn a $256 \times 512$ dictionary $\mathbf{A}$ and use it to initialize our D-LADMM. The network and parameter settings are the same as in the simulation experiments. Since the ground truth is unknown, we use the objective (\ref{D-LADMM_loss1}) as the loss function for training. The parameter in problem (\ref{SSC}) is set as $\lambda=0.5$, and Peak signal-to-noise ratio (PSNR) is used to evaluate the performance of various methods.
\par
Table~\ref{tab:PSNR} shows the comparison results at noise rate $10 \%$. For LADMM, which usually needs a large of iterations to converge, we examine its performance at $15$-th, $150$-th and $1,500$-th iterations---note that one iteration of LADMM corresponds to one block/layer of our D-LADMM. As can be seen, our $15$-layers D-LADMM achieves much higher PSNR than the LADMM at $15$-th iteration, and LADMM requires 1,500 iterations to obtain results comparable to our D-LADMM. In other words, compared with LADMM, D-LADMM achieves almost the same good solution within two order-of-magnitude fewer iterations. In Figure~\ref{fig:MONARCH}, we compare the visual quality of the denoised images by D-LADMM and LADMM, using the Lena image as the experimental data. It can be seen that the quality of the images recovered by D-LADMM is visibly higher than that of LADMM, especially for the areas full of rich details.
All the above results demonstrate the superiorities of our proposed D-LADMM over LADMM.
\vspace{-2mm}\subsection{Complexity Comparison}\vspace{-1mm}
Suppose that $\*{A} \in \mathbb{R}^{m \times d_1}$, $\*{B} \in \mathbb{R}^{m \times d_2}$  and $\*{X} \in \mathbb{R}^{m \times n}$, then the complexity of training D-LADMM is $\mathcal{O}((d_1+d_2)mnKp)$, where $K$ is the number of layers. Whereas the complexity of LADMM is $\mathcal{O}((d_1+d_2)mnt)$, where $t$ is the number of iterations. Usually, $t$ is one or two order-of-magnitude greater than $K$. Notice, that D-LADMM achieves comparable performance with LADMM only when $Kp<<t$, e.g., $t = 1,500$ and $Kp = 225$ as shown in Table~\ref{tab:PSNR}. In experiments, to both achieve an NMSE of $-13$dB with $n=10,000$ and $20,000$, D-LADMM needs $5$ and $9$ minutes (including training time), while LADMM needs $12$ and $22$ minutes, respectively. Therefore, even considering the training computational load, D-LADMM is still much faster than LADMM which needs a large iteration number. In addition, it is worth noting that, when a new data point comes, D-LADMM only needs a very low computational load for forward propagation.
\vspace{-2mm}
\section{Conclusion}\vspace{-1mm}
\label{sec:Conclusion}
In this paper, inspired by LADMM, we propose D-LADMM, a deep-learning-based optimization method for solving the constrained problem (\ref{Gene:L-ADMM}). Specificly, we first convert the proximal operator in LADMM to a special NN structure and replace the given matrix $\mathbf{A}$ in the proximal operator by learnable weights. Furthermore, we generalize the scalar to some elements-wise operations. From the theoretical perspective, we prove that, under some mild technical conditions, D-LADMM can obtain the linear convergence and converge faster than original LADMM. Experiments on simulative and real applications verify the superiority of D-LADMM.
\par
\section*{Acknowledgments}
The work of Guangcan Liu is supported in part by NSF of China under Grant 61622305 and Grant 61502238, in part by Natural Science Foundation of Jiangsu Province of China under Grant BK20160040, in part by SenseTime Research Fund. The work of Zhouchen Lin is supported in part by 973 Program of China under Grant 2015CB352502, in part by NSF of China under Grant 61625301 and Grant 61731018, in part by Qualcomm and Microsoft Research Asia.

\bibliography{sigproc}
\bibliographystyle{icml2019}


\newpage
\onecolumn
\appendix
\title{\textbf{Differentiable Linearized ADMM}\\(Supplementary Material)}
\date{\vspace{-5ex}}
\maketitle
\thispagestyle{empty} 
\icmlsetsymbol{equal}{*}

\begin{icmlauthorlist}
	\icmlauthor{Xingyu Xie}{equal,to}
	\icmlauthor{Jianlong Wu}{equal,to}
	\icmlauthor{Zhisheng Zhong}{to}
	\icmlauthor{Guangcan Liu}{ed}
	\icmlauthor{Zhouchen Lin}{to}
\end{icmlauthorlist}

\printAffiliationsAndNotice{\icmlEqualContribution} 

This Supplementary Material section contains the comparison with ADMM-Net, the technical proofs of convergence results, and some auxiliary lemmas of the manuscript entitled \emph{Differentiable Linearized ADMM}. It is structured as follows. 
Section~\ref{sec:a} presents the comparison between ADMM-Net and our proposed D-LADMM.
Section~\ref{Proof:D-LADMM} provides the proof of the lemmas and theorems in Section \ref{sec:D-LADMM}.
\section{Comparison with ADMM-Net}\label{sec:a}
\begin{table}[!ht]
	\centering
	\setlength{\tabcolsep}{20pt}
	\caption{PSNR comparison with ADMM-Net~\cite{sun2016deep} on image recovery task under different sampling ratios.}
	\begin{tabular}{|c|c|c|c|}
		\hline
		Sampling rate\textbackslash{}methods & Zero padding & ADMM-Net & D-LADMM(ours) \\ \hline
		20\%                   &    22.4          &     29.7     &       31.2                 \\ \hline
		30\%                                 &    24.7            &  31.8        &        33.5          \\ \hline
	\end{tabular}
	\label{tab:psnr_admmnet}
\end{table}
We also conduct a brain MR images recovery experiment to fairly compare with ADMM-Net~\cite{sun2016deep}.
Similar to \citet{sun2016deep}, we randomly
select $100$ images for training and $50$ images for testing.
We re-run these two methods on the same training and testing images. 
Table~\ref{tab:psnr_admmnet} shows the PSNR comparison.
We can see that our D-LADMM achieves better results than ADMM-Net under both two different sampling ratios.
Specifically, under $20\%$ sampling ratio, D-LADMM achieves a PSNR of $31.2$, which is better than the $29.7$ produced by ADMM-Net. As for the training time, D-LADMM costs $7.5$ hours on a CPU with $2.3$GHz, while ADMM-Net needs $17$ hours. So, D-LADMM is more efficient than ADMM-Net. The reason is as follows. For the compressed sensing problem, learnable parameters of D-LADMM are mostly the filters of $\*Z$-related iteration. In contrast, ADMM-Net has much more parameters to learn. With less learnable parameters, D-LADMM is more efficient on computation and memory. Most importantly, we provide the convergence analysis for learning-based method with constraints which is missing for ADMM-Net. In additional, we also provide a way to train the optimization inspired network in an unsupervised way in sharp contrast to the ADMM-Net's supervised training way.  
\section{Proofs for Section \ref{sec:D-LADMM}}
\label{Proof:D-LADMM}
\subsection{Proof of Lemma \ref{convexCond}}
\begin{proof}
Note that Lemma 3.1 in \cite{he2015non} inspired our proof.\par
From optimality conditions of minimization in the $\textbf{prox}_f(\cdot)$, we have:
\begin{equation}
f(\*Z) - f(\*Z_{k+1}) + \left\langle\*Z-\*Z_{k+1},\bm{\theta}_k \circ(\*{Z}_{k+1} - \*{Z}_k) +\*W_k^\top(\bm{\lambda}_k + \bm{\beta}_k\circ\*T_k)\right\rangle \geq 0, ~~\forall \*Z,\nonumber
\end{equation}
where $\*T_k =  \*A\*Z_k + \*E_k-\*X$.
By using the last line of (\ref{D-LADMMCov2}), we can get:
\begin{equation}\label{optZ}
f(\*Z) - f(\*Z_{k+1}) + \left\langle\*Z-\*Z_{k+1},\*W_k^\top\bm{\lambda}_{k+1} + \*W_k^\top\bm{\beta}_k\circ (\*E_k-\*E_{k+1}) + \#D_k(\*Z_{k+1}-\*Z_k)\right\rangle \geq 0.
\end{equation}
Similarly, from optimality conditions of $\*E_{k+1}$ in (\ref{D-LADMMCov2}), we have:
\begin{equation}
g(\*E) - g(\*E_{k+1}) + \left\langle\*E-\*E_{k+1},\bm{\lambda}_k + \bm{\beta}_k\circ(\*A\*Z_{k+1}+\*E_{k+1}-\*X) \right\rangle \geq 0, ~~\forall \*E.\nonumber
\end{equation}
Also, we can get:
\begin{equation}\label{optE}
g(\*E) - g(\*E_{k+1}) + \left\langle\*E-\*E_{k+1},\bm{\lambda}_{k+1}\right\rangle \geq 0, ~~\forall \*E.
\end{equation}
Note that, we have:
\begin{equation}\label{optLambda}
(\bm{\beta}_k)^{-1}\circ(\bm{\lambda}_{k+1} - \bm{\lambda}_k) -  (\*A\*Z_{k+1} + \*E_{k+1}-\*X) = 0.
\end{equation}
Combining (\ref{optZ}), (\ref{optE}) and (\ref{optLambda}) together, we get:
\begin{equation}
h(\*u) - h(\*u_{k+1}) + \left\langle\bm{\omega}-\bm{\omega}_{k+1}, \*N_k \right\rangle\geq 0,~~\forall \bm{\omega}_k,\nonumber
\end{equation}
where
\begin{equation}
\*N_k  =
\left(
  \begin{array}{c}
          \*W_k^\top(\bm{\lambda}_{k+1} + \bm{\beta}_k\circ(\*E_k-\*E_{k+1})) \\
          \bm{\lambda}_{k+1}\\
          \*{A}\*{Z}_{k+1}+\*{E}_{k+1} - \*{X}
 \end{array}
 \right)+
 \left(
   \begin{array}{c}
          \#D_k(\*Z_{k+1}-\*Z_k) \\
          0\\
          (\bm{\beta}_k)^{-1}\circ(\bm{\lambda}_k - \bm{\lambda}_{k+1})
 \end{array}
 \right).\nonumber
\end{equation}
Using the notations of $\#F_k, \#G_k$ and $\#H_k$, we complete the proof immediately.
\end{proof}
\subsection{Proof of Lemma \ref{Decent}}
\begin{proof}
Setting $\bm{\omega} = \bm{\omega}^*$ in (\ref{bound1}), we have:
\begin{equation}\label{bound2Right}
\left\langle\bm{\omega}_{k+1} - \bm{\omega}^*,\#H_k(\bm{\omega}_k - \bm{\omega}_{k+1})\right\rangle  \geq  h(\*u_{k+1}) - h(\*u^*)+\left\langle\bm{\omega}_{k+1} - \bm{\omega}^*,\#F_k(\bm{\omega}_{k+1})+\#G_k(\*E_k-\*E_{k+1})\right\rangle.
\end{equation}
Since $\bm{\omega}^* \in \*\Omega^*$ and by the Section 2.2 in~\cite{he2015non,liu2018partial}, we have:
\begin{equation}\label{F_Opt}
h(\*u_{k+1}) - h(\*u^*) + \left\langle\bm{\omega}_{k+1} - \bm{\omega}^*,\#F^*(\bm{\omega}^*)\right\rangle\geq 0,
\end{equation}
where
\begin{equation}
  \#F^*(\bm{\omega}) =
  \left(
  \begin{array}{c}
          \*A^\top\bm{\lambda} \\
          \bm{\lambda}\\
          \*{A}\*{Z}+\*{E} - \*{X}
 \end{array}
 \right).\nonumber
\end{equation}
According to the above inequality, we have:
\begin{equation}\label{Opt_F}
\begin{aligned}
h(\*u_{k+1}) - h(\*u^*) + \left\langle (\*\Delta_k)_{\bm{\omega}},~\#F_k(\bm{\omega}_{k+1})\right\rangle &= h(\*u_{k+1}) - h(\*u^*) +\left\langle(\*\Delta_k)_{\bm{\omega}},~\#F^*(\bm{\omega}^*)\right\rangle+\left\langle(\*\Delta_k)_{\bm{\omega}},
~\#F_k(\bm{\omega}_{k+1}) - \#F^*(\bm{\omega}^*)\right\rangle\\
&\geq \left\langle\bm{\lambda}_{k+1}, (\*W_k - \*A)(\*Z_{k+1} - \*Z^*)\right\rangle,
\end{aligned}
\end{equation}
where $(\*\Delta_k)_{\bm{\omega}} = \bm{\omega}_{k+1} - \bm{\omega}^*$.\\
Note that $\*A\*Z^*+\*E^* = \*X$ and by the notation of $\#G_k$ in Table \ref{tab:notations}, we obtain:
\begin{equation}\label{Opt_G}
\begin{aligned}
&\left\langle\bm{\omega}_{k+1} - \bm{\omega}^*,\#G_k(\*E_k-\*E_{k+1})\right\rangle\\
=&~\left\langle \bm{\beta}_k\circ(\*E_k-\*E_{k+1}),\*W_k\*Z_{k+1} + \*E_{k+1}-\*X\right\rangle - \left\langle \bm{\beta}_k\circ(\*E_k-\*E_{k+1}),(\*W_k-\*A)\*Z^*\right\rangle\\
=&~\left\langle (\*E_k-\*E_{k+1}),\bm{\beta}_k\circ(\*A\*Z_{k+1} + \*E_{k+1}-\*X)\right\rangle+ \left\langle (\*E_k-\*E_{k+1}),\bm{\beta}_k\circ\big((\*W_k-\*A)(\*Z_{k+1}-\*Z^*)\big)\right\rangle\\
=&~ \left\langle \bm{\beta}_k\circ(\*E_k-\*E_{k+1}),(\*W_k-\*A)(\*Z_{k+1}-\*Z^*)\right\rangle+ \left\langle (\*E_k-\*E_{k+1}),\bm{\lambda}_{k+1} - \bm{\lambda}_k\right\rangle
.
\end{aligned}
\end{equation}
From optimality conditions of $\*E_{k+1}$ in (\ref{D-LADMMCov2}), we have:
\begin{equation}
0 \in \partial g(\*E_{k+1}) + \bm{\lambda}_{k+1}.\nonumber
\end{equation}
Due to the convexity of the function $g(\cdot)$, we can conclude that:
$$\left\langle (\*E_k-\*E_{k+1}),\bm{\lambda}_{k+1} - \bm{\lambda}_k\right\rangle \geq 0.$$
Consider that $\*Z_k, \*E_k$ belong to the image of some proximal operators which are single-valued mappings. We prove finiteness by induction. We assume $\|\bm{\omega}_k\|_F<\infty$ for all $k\in[1,k-1]$. The single-valuedness of \textbf{prox}, finiteness  of $\bm{\omega}^{k}$  and closedness of the functions make $\bm{\omega}_{k+1}$ exist. Hence, when the $\sigma_k$ is small enough or even equal to $0$, then for any $(\*W_k, \bm{\theta}_k, \bm{\beta}_k) \in \mathcal{S}(\sigma_k, \*A)$, we can obtain:
\begin{equation}
\left\langle (\*E_k-\*E_{k+1}),\bm{\lambda}_{k+1} - \bm{\lambda}_k\right\rangle + \left\langle\bm{\lambda}_{k+1},\*\Delta_k (\*Z_{k+1} - \*Z^*) \right\rangle
+\left\langle \bm{\beta}_k\circ(\*E_k-\*E_{k+1}),\*\Delta_k(\*Z_{k+1}-\*Z^*)\right\rangle\geq 0\nonumber,
\end{equation}
where $\*\Delta_k = \*W_k-\*A$.
Combing the lower bounds in (\ref{Opt_F}) and (\ref{Opt_G}) together, we have the bound in (\ref{bound2}) immediately. Furthermore, due to (\ref{bound2}) we have:
\begin{equation}
\begin{aligned}
\|\bm{\omega}_k - \bm{\omega}^*\|_{\#H_k}^2 =& \|\bm{\omega}_k - \bm{\omega}_{k+1} + \bm{\omega}_{k+1}  - \bm{\omega}^*\|_{\#H_k}^2\\
=&\|\bm{\omega}_{k+1} - \bm{\omega}^*\|_{\#H_k}^2 + \|\bm{\omega}_k - \bm{\omega}_{k+1}\|_{\#H_k}^2 + 2\langle\bm{\omega}_{k+1} - \bm{\omega}^*, \#H_k(\bm{\omega}_k - \bm{\omega}_{k+1})\rangle\\
\geq& \|\bm{\omega}_{k+1} - \bm{\omega}^*\|_{\#H_k}^2 + \|\bm{\omega}_k - \bm{\omega}_{k+1}\|_{\#H_k}^2.\nonumber
\end{aligned}
\end{equation}
We get the bound in $(\ref{bound3})$ directly and can easily derive the finiteness of $\|\bm{\omega}_k\|_F$ from it. We finish the proof.
\end{proof}

\subsection{Proof of Theorem \ref{main_thm1}}
\begin{proof}
From the Lemma \ref{Decent}, given one $\bm{\omega}^*\in \*\Omega^*$, there exist proper $(\*W_k, \bm{\theta}_k, \bm{\beta}_k) \in \mathcal{S}(\sigma_k, \*A)$ such that:
\begin{equation}
\begin{aligned}
&\sum_{k=0}^{\infty}\|\bm{\omega}_k - \bm{\omega}_{k+1}\|_{\#H_k}^2 \leq \sum_{k=0}^{\infty}\|\bm{\omega}_k - \bm{\omega}^*\|_{\#H_k}^2 - \|\bm{\omega}_{k+1} - \bm{\omega}^*\|_{\#H_k}^2\\
& \leq  \|\bm{\omega}^0 - \bm{\omega}^*\|_{\#H^0}^2 + \sum_{k=0}^{\infty}\left|\|\bm{\omega}_{k+1} - \bm{\omega}^*\|_{(\#H_{k+1} - \#H_k)}^2\right|.\nonumber
\end{aligned}
\end{equation}
If we define some large enough $\bm{\theta}^*, \bm{\beta}^*$ and let $\sigma_k\rightarrow 0$, $\bm{\theta}_k \rightarrow\bm{\theta}^*$ and $\bm{\beta}_k\rightarrow\bm{\beta}^*$ with a speed faster than $1/k^2$, i.e., $\|\#H_{k+1} - \#H_k\| = O(1/k^2)$, then we can get $\sum_{k=0}^{\infty}\left|\|\bm{\omega}_{k+1} - \bm{\omega}^*\|_{(\#H_{k+1} - \#H_k)}^2\right| < \infty$, and thus:
$$\sum_{k=0}^{\infty}\|\bm{\omega}_k - \bm{\omega}_{k+1}\|_{\#H_k}^2 < \infty.$$
Consequently, we have $\|\bm{\omega}_k - \bm{\omega}_{k+1}\|_{\#H_k}^2\rightarrow 0$.
Since $(\*W_k, \bm{\theta}_k, \bm{\beta}_k) \in \mathcal{S}(\sigma_k, \*A)$, we have $\#H_k \succ0$. Then,
by (\ref{bound3}) and $\sum_{k=0}^{\infty}\|\bm{\omega}_{k+1} - \bm{\omega}^*\|_{(\#H_{k+1} - \#H_k)}^2 < \infty$, we know that the sequence $\{\bm{\omega}_k\}$ is bounded and let $\{\bm{\omega}^{k_t}\}$ be a subsequence of $\{\bm{\omega}_k\}$ converging to $\bm{\omega}^\infty$. Considering the inequality (\ref{bound1}) for the subsequence $\{\bm{\omega}^{k_t}\}$, taking the limit over $t$. Since we let $\sigma_k\rightarrow 0$, then use the fact that $\|\bm{\omega}^{k_t} - \bm{\omega}^{k_{t+1}}\|_{\#H_k}^2\rightarrow 0$, we obtain:
\begin{equation}
h(\*u) - h(\*u_\infty) + \left\langle\bm{\omega} - \bm{\omega}_\infty,\#F_{\infty}(\bm{\omega}_\infty)\right\rangle\geq 0,\nonumber
\end{equation}
where $\#F^{\infty}(\cdot) = \#F^*(\cdot)$ define in (\ref{F_Opt}) when $\sigma_k\rightarrow 0$. We conclude that $\bm{\omega}^\infty \in \*\Omega^*$. Since $\#H_k \succ0$ for all $k$ and $\|\bm{\omega}_k - \bm{\omega}_{k+1}\|_{\#H_k}^2\rightarrow 0$, we immediately have $\bm{\omega}_k\rightarrow \bm{\omega}_\infty$
as $k \rightarrow \infty$ and the proof is complete.
\end{proof}

\subsection{Proof of Lemma \ref{non-zero}}
\begin{proof}
On the one hand any  $(\*W_k, \bm{\theta}_k, \bm{\beta}_k) \in \mathcal{S}(\sigma, \*A)$ does not change the solution of (\ref{D-LADMMCov2}), on the other hand \textbf{prox} is single-valued. Hence we conclude that $( \*A\*Z_k + \*E_k-\*X) = 0$, and $\bm{\lambda}_k = 0$ or $\sigma_k = 0$.
According to the optimality conditions of the (\ref{D-LADMMCov2}) and $\bm{\omega}_k = \bm{\omega}_{k+1}$, we have:
\begin{equation}
 0 \in \partial f(\*Z_{k+1}) + \*A^T\bm{\lambda}_{k+1},~~~\*A\*Z_{k+1} + \*E_{k+1}-\*X = 0.\nonumber
\end{equation}
Note that we already have $0 \in \partial g(\*E_{k+1}) + \bm{\lambda}_{k+1}$. Hence, $\bm{\omega}_k = \bm{\omega}_{k+1}$ are the KKT point. Since the original problem (\ref{Gene:L-ADMM}) is convex with linear constraint, the KKT conditions are also sufficient, thus we conclude that $\bm{\omega}_k = \bm{\omega}_{k+1} \in \*\Omega^*$.
\end{proof}

\subsection{Proof of Theorem \ref{main_thm2}}
\begin{proof}
We assume that there exists some $(\*W_k, \bm{\theta}_k, \bm{\beta}_k) \in \mathcal{S}(\sigma, \*A)$ to make the $\bm{\omega}_{k+1}\neq\bm{\omega}_k$, or by the Lemma \ref{non-zero} we have $\bm{\omega}_{k+1} \in \*\Omega^*$ and finish the proof.\\
Without loss of generality, we assume that $\|\bm{\omega}_{k+1} - \bm{\omega}_k\|_{\#H_k}^2\neq0$, otherwise we can perturb $(\*W_k, \bm{\theta}_k, \bm{\beta}_k) \in \mathcal{S}(\sigma_k, \*A)$ to make $\bm{\omega}_{k+1} \neq \bm{\omega}_k$. Due to $\|\bm{\omega}_{k+1} - \bm{\omega}_k\|_{\#H_k}^2\neq0$, there exists $\kappa_k > 0$ such that:
\begin{equation}\label{FEB}
\operatorname{dist}_{\#H_k}^2(\bm{\omega}_{k+1}, \*\Omega^*)\leq \kappa_k \|\bm{\omega}_{k+1} - \bm{\omega}_k\|_{\#H_k}^2.
\end{equation}
Following from (\ref{bound3}), we have:
\begin{equation}
\operatorname{dist}_{\#H_k}^2(\bm{\omega}_{k+1}, \*\Omega^*)\leq  \operatorname{dist}_{\#H_k}^2(\bm{\omega}_k, \*\Omega^*)- \|\bm{\omega}_k - \bm{\omega}_{k+1}\|_{\#H_k}^2.\nonumber
\end{equation}
Combing the above inequality with (\ref{FEB}), we get:
\begin{equation}
\operatorname{dist}_{\#H_k}^2(\bm{\omega}_{k+1}, \*\Omega^*)\leq \left(1+ \frac{1}{\kappa_k}\right)^{-1}  \operatorname{dist}_{\#H_k}^2(\bm{\omega}_k, \*\Omega^*).\nonumber
\end{equation}
For the $\bm{\omega}^* \in \*\Omega^*$ such that $\operatorname{dist}_{\#H_k}^2(\bm{\omega}_{k+1}, \*\Omega^*) = \|\bm{\omega}_{k+1} - \bm{\omega}^*\|_{\#H_k}^2$, we have:
\begin{equation}
\operatorname{dist}_{\#H_{k+1}}^2(\bm{\omega}_{k+1}, \*\Omega^*)\leq \|\*\Delta_k\|_{\#H_k}^2 + \|\*\Delta_k\|_{\#H_{k+1} - \#H_k}^2,\nonumber
\end{equation}
where $\*\Delta_k = \bm{\omega}_{k+1} - \bm{\omega}^*$.\\
If we define some large enough $\bm{\theta}^*, \bm{\beta}^*$ and let $\sigma_k\rightarrow 0$, $\bm{\theta}_k \rightarrow\bm{\theta}^*$ and $\bm{\beta}_k\rightarrow\bm{\beta}^*$ with a fast enough speed, i.e., $\|\#H_{k+1} - \#H_k\|_2 = O(1/\alpha_k)$ where $\alpha>1$. We can obtain:
\begin{equation}
\begin{aligned}
\operatorname{dist}_{\#H_{k+1}}^2(\bm{\omega}_{k+1}, \*\Omega^*)\leq (1+\frac{1}{c\alpha_k})\operatorname{dist}_{\#H_k}^2(\bm{\omega}_{k+1}, \*\Omega^*),\nonumber
\end{aligned}
\end{equation}
where $c>0$ is some constant. When $k$ is large enough, with large enough $c$ and $\alpha$, we can conclude that:
$$\left(1+\frac{1}{c\alpha_k}\right)\left(1+ \frac{1}{\kappa_k}\right)^{-1}<1.$$
Hence, we have $\operatorname{dist}_{\#H_{k+1}}^2(\bm{\omega}_{k+1}, \*\Omega^*)< \operatorname{dist}_{\#H_k}^2(\bm{\omega}_k, \*\Omega^*)$, and finish the proof.
\end{proof}

\subsection{Proof of Theorem \ref{main_thm3}}
\begin{proof}
\textbf{Case 1:} When $k\geq K_0$, for one $\bm{\omega}^* \in \*\Omega^*$ such that $\operatorname{dist}_{\#H^*}^2(\widetilde{\bm{\omega}}, \*\Omega^*) = \|\widetilde{\bm{\omega}} - \bm{\omega}^*\|_{\#H^*}^2$, we have:
\begin{equation}
\operatorname{dist}_{\#H_{k+1}}^2(\widetilde{\bm{\omega}}, \*\Omega^*)\leq \operatorname{dist}_{\#H^*}^2(\widetilde{\bm{\omega}}, \*\Omega^*) + \left|\|\widetilde{\bm{\omega}} - \bm{\omega}^*\|_{\#H_{k+1}-\#H^*}^2\right|.\nonumber
\end{equation}
If we let $\sigma_k\rightarrow 0$, $\bm{\theta}_k \rightarrow\bm{\theta}^*$ and $\bm{\beta}_k\rightarrow\bm{\beta}^*$ with a linear decay rate , i.e., $\|\#H_{k+1} - \#H_k\| = O(1/\alpha_k)$ where $\alpha>1$. Then we have:
\begin{equation}\label{leftBound}
\operatorname{dist}_{\#H_{k+1}}^2(\widetilde{\bm{\omega}}, \*\Omega^*)< 2~\operatorname{dist}_{\#H^*}^2(\widetilde{\bm{\omega}}, \*\Omega^*).
\end{equation}
Similarly, we have:
\begin{equation}\label{RightBound}
\|\widetilde{\bm{\omega}} - \bm{\omega}_k\|_{\#H^*}^2 = \|\widetilde{\bm{\omega}} - \bm{\omega}^*\|_{\#H_k}^2+\left|\|\widetilde{\bm{\omega}} - \bm{\omega}^*\|_{\#H^*-\#H_k}^2\right|<2\|\widetilde{\bm{\omega}} - \bm{\omega}^*\|_{\#H_k}^2.
\end{equation}
Combing (\ref{leftBound}) and (\ref{RightBound}) together, we have:
\begin{equation}\label{ErrBound2}
\operatorname{dist}_{\#H_{k+1}}^2(\widetilde{\bm{\omega}}, \*\Omega^*)< \frac{\kappa}{4} \|\widetilde{\bm{\omega}} - \bm{\omega}_k\|_{\#H_k}^2.
\end{equation}
Note that $\|\cdot\|_{\#H}$ is a norm if $\#H\succ0$, then we have:
\begin{equation}
\begin{aligned}\label{deltaBound}
& \operatorname{dist}_{\#H_{k+1}}(\bm{\omega}_{k+1}, \*\Omega^*)\leq \|\bm{\omega}_{k+1} - \widetilde{\bm{\omega}}\|_{\#H_{k+1}}+\operatorname{dist}_{\#H_{k+1}}(\widetilde{\bm{\omega}}, \*\Omega^*)\\
<&\frac{\sqrt{\kappa}}{2} (\|\bm{\omega}_{k+1} - \bm{\omega}_k\|_{\#H_k}+ \|\bm{\omega}_{k+1} - \widetilde{\bm{\omega}}\|_{\#H_k}) + \|\bm{\omega}_{k+1} - \widetilde{\bm{\omega}}\|_{\#H_{k+1}}.
\end{aligned}
\end{equation}
We already have $\widetilde{\bm{\omega}} = \#T(\*A, \bm{\theta}^*, \bm{\beta}^*)(\bm{\omega}_k)$ and $\bm{\omega}_{k+1} = \#T(\*W_k, \bm{\theta}_k, \bm{\beta}_k)(\bm{\omega}_k)$. The operator $\#T$ consists of \textbf{prox} and linear operators, hence $\#T$ is firmly-nonexpansive with respect to the parameter $(\*W, \bm{\theta}, \bm{\beta})$, i.e.,:
\begin{equation}
\|\bm{\omega}_{k+1} - \widetilde{\bm{\omega}}\|_F\leq O(\|\*\Delta_{\Theta}\|_F),\nonumber
\end{equation}
where $\*\Delta_{\Theta} = (\*W_k, \bm{\theta}_k, \bm{\beta}_k) - (\*A, \bm{\theta}^*, \bm{\beta}^*)$. Due to $\#H_k\succ0$ for all $k$, we can obtain:
\begin{equation}\label{deltaBound2}
\frac{\sqrt{\kappa}}{2} \|\bm{\omega}_{k+1} - \widetilde{\bm{\omega}}\|_{\#H_k} + \|\bm{\omega}_{k+1} - \widetilde{\bm{\omega}}\|_{\#H_{k+1}}\leq O(\|\*\Delta_{\Theta}\|_F).
\end{equation}
Since $(\*W_k, \bm{\theta}_k, \bm{\beta}_k) \rightarrow (\*A, \bm{\theta}^*, \bm{\beta}^*)$ with a linear decay rate, without loss of generality, we let $\{(\*W_k, \bm{\theta}_k, \bm{\beta}_k)\}$ convergence faster than $\{\bm{\omega}_k\}$.
With large enough $K_0$ and proper $c, \alpha$, we have:
\begin{equation}\label{deltaBound3}
O(\|\*\Delta_{\Theta}\|_F) \leq \frac{1}{c\alpha_k} \leq \frac{1}{2}\operatorname{dist}_{\#H_{k+1}}(\bm{\omega}_{k+1}, \*\Omega^*).
\end{equation}
Combing the inequalities (\ref{deltaBound})-(\ref{deltaBound3}), we conclude that:
\begin{equation}\label{ErrBound2}
\operatorname{dist}_{\#H_{k+1}}^2(\bm{\omega}_{k+1}, \*\Omega^*)< \kappa \|\bm{\omega}_{k+1} - \bm{\omega}_k\|_{\#H_k}^2.
\end{equation}
\textbf{Case 2:} When $k< K_0$, from the convergence of D-LADMM in Theorem \ref{main_thm1} and the inequality (\ref{bound3}), we know that $\operatorname{dist}_{\#H_{k+1}}^2(\bm{\omega}_{k+1}, \*\Omega^*)<\|\bm{\omega}^0 - \bm{\omega}^*\|_{\#H^0}^2 + \sum_{k=0}^{K_0}\left|\|\bm{\omega}_{k+1} - \bm{\omega}^*\|_{(\#H_{k+1} - \#H_k)}^2\right|<\infty$.
Hence there exists one constant $B>0$ such that $\operatorname{dist}_{\#H_{k+1}}^2(\bm{\omega}_{k+1}, \*\Omega^*)<B$. Since $\|\bm{\omega}_k - \bm{\omega}_{k+1}\|_{\#H_k}^2\neq 0$, there exists one constant $\epsilon>0$ such that $\|\bm{\omega}_k - \bm{\omega}_{k+1}\|_{\#H_k}^2 > \epsilon$ when $k< K_0$. We immediately
have:
\begin{equation}
\operatorname{dist}_{\#H_{k+1}}^2(\bm{\omega}_{k+1}, \*\Omega^*)< \frac{B}{\epsilon}\|\bm{\omega}_{k+1} - \bm{\omega}_k\|_{\#H_k}^2.\nonumber
\end{equation}
Letting $\widetilde{\kappa} = \max\{\frac{B}{\epsilon}, \kappa\}$. Following from (\ref{bound3}) and the Monotonicity of D-LADMM in Theorem \ref{main_thm2}, we get
\begin{equation}
\operatorname{dist}_{\#H_{k+1}}^2(\bm{\omega}_{k+1}, \*\Omega^*)< \left(1+ \frac{1}{\widetilde{\kappa}}\right)^{-1}  \operatorname{dist}_{\#H_k}^2(\bm{\omega}_k, \*\Omega^*).\nonumber
\end{equation}
We finish the proof.
\end{proof}

\subsection{Proof of Lemma \ref{lem:weakercond}}
\begin{proof}
From the monotonicity of D-LADMM in Theorem \ref{main_thm2}, we have $dist(\bm{\omega}, \*\Omega^*)\leq \epsilon$ for some constant $\epsilon$ when $k$ large enough, then the error bound condition (\ref{ErrBound}) can be rewritten as:
\begin{equation}
\operatorname{dist}_{\#H^*}^2(\widetilde{\bm{\omega}}, \*\Omega^*)\leq \frac{\kappa}{16} \|\widetilde{\bm{\omega}} - \bm{\omega}_k\|_{\#H^*}^2 ~\forall \bm{\omega}_k,
\end{equation}
where $dist(\bm{\omega}, \*\Omega^*)\leq \epsilon$. The left proof is the same as that in the Section 3.2 of~\cite{liu2018partial}.
\end{proof}
\subsection{Proof of Lemma \ref{D-LADMM:example}}
\begin{proof}
In this proof, we only consider the case for $n=1$, namely $\*Z\in \mathbb{R}^{d_1}$ and $\*E\in \mathbb{R}^{d_2}$, in the following, we use the lowercase letters  to denote the vector. The result for the matrix case can be derived from the vector case. Without loss of generality, we let $\bm{\theta}^* = \*1$ and $\bm{\beta}^* = \*1$, then by the condition (\ref{PGD:Assumption}), we get $\|\*A\|<1$ and $\|\*B\|<1$.
\par 
When fixing $((\*W_1)_k, (\*W_2)_k, \bm{\theta}_k, \bm{\beta}_k)$ to $(\*A, \*B, \bm{\theta}^*, \bm{\beta}^*)$, D-LADMM degenerate to LADMM:
\[
\left\{
\begin{aligned}
& \*{z}_{k+1} = \textbf{prox}_{f}\left\{\*{z}_k - \*A^\top(\bm{\lambda}_k + \*A\*z_k + \*B\*e_k-\*x )\right\},\\
& \*e_{k+1} = \textbf{prox}_{g}\left\{\*{e}_k - \*B^\top(\bm{\lambda}_k + \*A\*z_{k+1} + \*B\*e_k-\*x )\right\},\\
& \bm{\lambda}_{k+1} = \bm{\lambda}_k + \*A\*z_{k+1} + \*B\*e_{k+1}-\*x,
\end{aligned}
\right.
\]
For convenience, we also let $\bm{\theta}_k = \*1$, $\bm{\beta}_k = \*1$ and $(\*W_1)_k = \*A$ in the proximal operator and left one learnable scalar for the Lagrange multiplier, then our D-LADMM reads as:
\[
\left\{
\begin{aligned}
& \*{z}_{k+1} = \textbf{prox}_{f}\left\{\*{z}_k - \*A^\top(\bm{\lambda}_k + \*A\*z_k + \*B\*e_k-\*x )\right\},\\
& \*e_{k+1} = \textbf{prox}_{g}\left\{\*{e}_k - \*W_k^\top(\bm{\lambda}_k + \*A\*z_{k+1} + \*B\*e_k-\*x )\right\},\\
& \bm{\lambda}_{k+1} = \bm{\lambda}_k + \*A\*z_{k+1} + \*B\*e_{k+1}-\*x,
\end{aligned}
\right.
\]
We denote $\textbf{prox}_{g}(\cdot)$ as $\eta_g(\cdot)$. Given $(\*z_k,\*e_k,\bm{\lambda}_k)$, we let $\*y_k = \bm{\lambda}_k + \*A\*z_{k+1} + \*B\*e_k-\*x$. Now we consider the update of the variable $\*e$, if we have:
\begin{equation}\label{example:pair}
\eta_g(\*{e}_k - \*W_k^\top\*y_k) - \eta_f(\*{e}_k - \*B^\top\*y_k) = c (\*e^* - \eta_f(\*{e}_k - \*B^\top\*y_k)),
\end{equation}
for some constant $0<c<1$. The above equation implies the output of D-LADMM is closer to the solution set than LADMM on variable $\*e$. Actually, the term on the right side of the equation (\ref{example:pair}) is the direction from the output of LADMM to the solution set. Due to the bijectiveness of the $\eta_g(\cdot)$, the equation (\ref{example:pair}) can be rewritten as
\[
\*\Delta_w^\top\*y_k = \eta^{-1}_g\left(c (\*e^* - \eta_f(\*{e}_k - \*B^\top\*y_k))\right)\triangleq \*r_k.
\]
From the definition of $\*y_k$, we know that $\*y_k$ belongs to the space span by the matrix $\*A$ and $\*B$. We denote the basis of the the space span by the matrix $\*A$ and $\*B$ as $\*M \in \mathbb{R}^{m\times d_3}$. $\*y_k$ can be written as $\*y_k = \*M\*a_k$, where $\*a_k \in \mathbb{R}^{d_3}$. Now we define the support set of $\*r_k$ as $\mathcal{S}_r = \operatorname{support}(\*r_k)$ and let the support set of $\*a_k$ as $\mathcal{S}_a = \operatorname{support}(\*a_k)$. Given any $i\in [d_1]$, we consider cases.
\par
\textbf{Case 1:} $i \not\in  \mathcal{S}_r$, then $(\*r_k) = 0$. Under this circumstance, we let the $i$-th column of $\*\Delta_w$ as zero column, i.e., $(\*\Delta_w)_i = \*0$.\\
\textbf{Case 2:} $i \in  \mathcal{S}_r$, then $(\*r_k)_i \neq 0$. We can set the $i$-th column of $\*\Delta_w$ as $t\*M_j$, where $j\in \mathcal{S}_a$ and $t = (\*r_k)_i/ (\*a_k)_j$. Then we have $(\*\Delta_w)_i^\top\*M\*a_k = (\*a_k)_j \frac{(\*r_k)_i }{(\*a_k)_j} = (\*r_k)_i$.\\
By the above way of setting, we find a $\*\Delta_w$ such that having (\ref{example:pair}) satisfied. Let the output of the D-LADMM be $\widehat{\bm{\omega}}_{k+1}$ and the output of the LADMM be $\bm{\omega}_{k+1}$, then
\begin{equation}\label{example:bound}
\|\bm{\omega}_{k+1} - \bm{\omega}^*\|_2^2 - \|\widehat{\bm{\omega}}_{k+1} - \bm{\omega}^*\|_2^2 \geq c\|\Delta_e\|_2^2 - c\|\*B\Delta_e\|_2^2 \geq c (1-\|B\|_2)\|\Delta_e\|_2^2 > 0,
\end{equation}
where $\Delta_e = \*e^* - \eta_f(\*{e}_k - \*B^\top\*y_k)$. Note that $\bm{\omega}_{k+1} \not\in \*\Omega^*$, then $\|\Delta_e\|_2>0$. The last inequality comes from the condition $\|\*B\|<1$.\\
The above (\ref{example:bound}) implies the output of D-LADMM is closer to the solution set than LADMM. We finish the proof.
\end{proof}


\end{document}